%% file: iclr2023_conference.tex
\documentclass{article} 
\usepackage{iclr2023_conference,times}

\input{math_commands.tex}

\usepackage{hyperref}
\usepackage{url}
\usepackage{multirow}
\usepackage{graphicx}
\usepackage{caption}
\usepackage{booktabs}
\usepackage{makecell}
\usepackage{subfigure}
\usepackage{floatrow}
\newfloatcommand{capbtabbox}{table}[][\FBwidth]
\usepackage{algorithmicx}

\include{Sections/mlVecMat}

\title{Large Language Models Can Self-Improve}

\author{Jiaxin Huang${}^1$\thanks{Work was done during Google internship.} \quad Shixiang Shane Gu${}^2$ \quad Le Hou${}^2$\thanks{Corresponding author.} \quad Yuexin Wu${}^2$ \quad Xuezhi Wang${}^2$ \quad \\ 
\textbf{Hongkun Yu${}^2$ \quad Jiawei Han${}^1$}\\
${}^1$University of Illinois at Urbana-Champaign \ \ \ ${}^2$Google \\
${}^1$\texttt{\{jiaxinh3, hanj\}@illinois.edu} \ \ \ ${}^2$\texttt{\{shanegu, lehou, crickwu,}\\\texttt{ xuezhiw, hongkuny\}@google.com}
}

\newcommand{\method}{\textbf{LMSI}\xspace}
\newcommand{\bigmodel}{PaLM-540B}
\newcommand{\midmodel}{PaLM-62B}
\newcommand{\smallmodel}{PaLM-8B}
\renewcommand{\headrulewidth}{0pt}

\iclrfinalcopy 
\begin{document}

\maketitle

\input{Sections/0-abs}

\input{Sections/1-intro}

\input{Sections/2-related}

\input{Sections/3-method}

\input{Sections/4-setup}

\input{Sections/5-exp}

\input{Sections/6-concl}

\input{Sections/ack}



\bibliography{iclr2023_conference}
\bibliographystyle{iclr2023_conference}

\input{Sections/7-app}

\end{document}

%% file: math_commands.tex
\usepackage{amsmath,amsfonts,bm}

\def\eqref#1{equation~\ref{#1}}

\def\1{\bm{1}}

\DeclareMathAlphabet{\mathsfit}{\encodingdefault}{\sfdefault}{m}{sl}
\SetMathAlphabet{\mathsfit}{bold}{\encodingdefault}{\sfdefault}{bx}{n}

\DeclareMathOperator*{\argmax}{arg\,max}

%% file: Sections/mlVecMat.tex
\usepackage{amsmath}
\usepackage{amsfonts}
\usepackage{amssymb}
\usepackage{wrapfig}
\usepackage{multirow}
 \usepackage{mathtools} 

\usepackage{verbatim}

\usepackage{anyfontsize}

\usepackage{color}
\usepackage{tikz}
\usetikzlibrary{arrows,shapes,snakes,automata,backgrounds,fit,petri}
\usepackage{adjustbox}

\newcommand{\RN}[1]{%
	\textup{\lowercase\expandafter{\it \romannumeral#1}}%
}
\usepackage{soul}
\usepackage{tikz}
\usetikzlibrary{calc}
\usetikzlibrary{decorations.pathmorphing}

\makeatletter

\newcommand{\defhighlighter}[3][]{%
  \tikzset{every highlighter/.style={color=#2, fill opacity=#3, #1}}%
}

\defhighlighter{yellow}{.5}

\newcommand{\highlight@DoHighlight}{
  \fill [ decoration = {random steps, amplitude=1pt, segment length=15pt}
        , outer sep = -15pt, inner sep = 0pt, decorate
        , every highlighter, this highlighter ]
        ($(begin highlight)+(0,8pt)$) rectangle ($(end highlight)+(0,-3pt)$) ;
}

\newcommand{\highlight@BeginHighlight}{
  \coordinate (begin highlight) at (0,0) ;
}

\newcommand{\highlight@EndHighlight}{
  \coordinate (end highlight) at (0,0) ;
}

\newdimen\highlight@previous
\newdimen\highlight@current

\DeclareRobustCommand*\highlight[1][]{%
  \tikzset{this highlighter/.style={#1}}%
  \SOUL@setup
  \def\SOUL@preamble{%
    \begin{tikzpicture}[overlay, remember picture]
      \highlight@BeginHighlight
      \highlight@EndHighlight
    \end{tikzpicture}%
  }%
  \def\SOUL@postamble{%
    \begin{tikzpicture}[overlay, remember picture]
      \highlight@EndHighlight
      \highlight@DoHighlight
    \end{tikzpicture}%
  }%
  \def\SOUL@everyhyphen{%
    \discretionary{%
      \SOUL@setkern\SOUL@hyphkern
      \SOUL@sethyphenchar
      \tikz[overlay, remember picture] \highlight@EndHighlight ;%
    }{%
    }{%
      \SOUL@setkern\SOUL@charkern
    }%
  }%
  \def\SOUL@everyexhyphen##1{%
    \SOUL@setkern\SOUL@hyphkern
    \hbox{##1}%
    \discretionary{%
      \tikz[overlay, remember picture] \highlight@EndHighlight ;%
    }{%
    }{%
      \SOUL@setkern\SOUL@charkern
    }%
  }%
  \def\SOUL@everysyllable{%
    \begin{tikzpicture}[overlay, remember picture]
      \path let \p0 = (begin highlight), \p1 = (0,0) in \pgfextra
        \global\highlight@previous=\y0
        \global\highlight@current =\y1
      \endpgfextra (0,0) ;
      \ifdim\highlight@current < \highlight@previous
        \highlight@DoHighlight
        \highlight@BeginHighlight
      \fi
    \end{tikzpicture}%
    \the\SOUL@syllable
    \tikz[overlay, remember picture] \highlight@EndHighlight ;%
  }%
  \SOUL@
}
\makeatother

\usepackage{booktabs}

\usepackage{tcolorbox}

\makeatletter
\newcommand{\distas}[1]{\mathbin{\overset{#1}{\kern\z@\sim}}}%

\usepackage{enumitem}

\usepackage[lined,boxed,commentsnumbered,ruled,linesnumbered]{algorithm2e}

\newcommand{\beq}{\vspace{0mm}\begin{equation}}
\newcommand{\eeq}{\vspace{0mm}\end{equation}}
\newcommand{\beqs}{\vspace{0mm}\begin{eqnarray}}
\newcommand{\eeqs}{\vspace{0mm}\end{eqnarray}}
\newcommand{\barr}{\begin{array}}
\newcommand{\earr}{\end{array}}

\newcommand{\rvec}{\boldsymbol{r}}

\usepackage[normalem]{ulem}

\newcommand{\Dcal}{\mathcal{D}}

\ifx\assumption\undefined
\newtheorem{assumption}{Assumption}
\fi

\ifx\definition\undefined
\newtheorem{definition}{Definition}
\fi

\ifx\remark\undefined
\newtheorem{remark}{Remark}
\fi

\usepackage{tabularx,caption}
\usepackage[leftcaption]{sidecap}

\usepackage{booktabs}
\usepackage{floatrow}


%% file: Sections/0-abs.tex
\begin{abstract}
Large Language Models (LLMs) have achieved excellent performances in various tasks. However, fine-tuning an LLM requires extensive supervision. Human, on the other hand, may improve their reasoning abilities by self-thinking without external inputs. In this work, we demonstrate that an LLM is also capable of self-improving with only unlabeled datasets.
We use a pre-trained LLM to generate ``high-confidence'' rationale-augmented answers for unlabeled questions using Chain-of-Thought prompting and self-consistency, and fine-tune the LLM using those self-generated solutions as target outputs. We show that our approach improves the general reasoning ability of a 540B-parameter LLM (74.4\%$\rightarrow$82.1\% on GSM8K, 78.2\%$\rightarrow$83.0\% on DROP, 90.0\%$\rightarrow$94.4\% on OpenBookQA, and 63.4\%$\rightarrow$67.9\% on ANLI-A3) and achieves state-of-the-art-level performance, without any ground truth label. We conduct ablation studies and show that fine-tuning on reasoning is critical for self-improvement.
\end{abstract}

%% file: Sections/1-intro.tex
\section{Introduction}

Scaling has enabled Large Language Models (LLMs) to achieve state-of-the-art performance on a range of Natural Language Processing (NLP) tasks~\citep{Wang2018GLUEAM, Wang2019SuperGLUEAS, Rajpurkar2016SQuAD1Q}. More importantly, new capabilities have emerged from LLMs as they are scaled to hundreds of billions of parameters~\citep{wei2022emergent}: in-context few-shot learning \citep{Brown2020LanguageMA} makes it possible for an LLM to perform well on a task it never trained on with only a handful of examples; Chain-of-Thought (CoT) prompting \citep{Wei2022ChainOT, step_by_step} demonstrates strong reasoning ability of LLMs across diverse tasks with or without few-shot examples; self-consistency \citep{Wang2022SelfConsistencyIC} further improves the performance via self-evaluating multiple reasoning paths.

Despite these incredible capabilities of models trained on large text corpus~\citep{Brown2020LanguageMA,Chowdhery2022PaLMSL}, fundamentally improving the model performances beyond few-shot baselines still requires finetuning on an extensive amount of \textit{high-quality supervised} datasets. FLAN~\citep{wei2021finetuned,chung2022scaling} and T0~\citep{sanh2021multitask} curated tens of benchmark NLP datasets to boost zero-shot task performances on unseen tasks; InstructGPT~\citep{ouyang2022training} crowd-sourced many human answers for diverse sets of text instructions to better align their model to human instructions. 
While significant efforts were committed on collecting high-quality supervised datasets, human brain, on the contrary, is capable of the metacognition process \citep{dunlosky2008metacognition}, where we can refine our own reasoning ability without external inputs. 

In this paper, we study how an LLM is able to \textit{self-improve} its reasoning ability without supervised data. We show that using only
input sequences (without ground truth output sequences) from multiple NLP task datasets, a pre-trained LLM is able to improve performances for both in-domain and out-of-domain tasks. 
Our method is shown in Figure~\ref{fig:framework}: we first sample multiple predictions using few-shot Chain-of-Thought (CoT)~\citep{Wei2022ChainOT} as prompts, filter ``high-confidence'' predictions using majority voting~\citep{Wang2022SelfConsistencyIC}, and finally finetune the LLM on these high-confidence predictions. The resulting model shows improved reasoning in both greedy and multi-path evaluations.
We call the model fine-tuned in this way as \textbf{Language Model Self-Improved (LMSI)}. This is similar to how a human brain sometimes learns: given a question, think multiple times to derive different possible results, conclude on how the question should be solved, and then learn from or memorize its own solution.
We empirically verify our method using a pre-trained \bigmodel~LLM, where our method not only improves training task performances (74.4\%$\rightarrow$82.1\% on GSM8K, 78.2\%$\rightarrow$83.0\% on DROP, 90.0\%$\rightarrow$94.4\% on OpenBookQA, and 63.4\%$\rightarrow$67.9\% on ANLI-A3), but also enhances out-of-domain (OOD) test tasks (AQUA, 
StrategyQA, 
MNLI), achieving state-of-the-art performances in many tasks without relying on supervised ground truth answers. Lastly, we conduct preliminary studies on self-generating additional input questions and few-shot CoT prompts, which could further reduce the amount of human effort required for model self-improving, and ablation studies on important hyperparameters of our approach. 
We hope our simple approach and strong empirical results could encourage more future work by the community to investigate optimal performances of pretrained LLMs without additional human supervision.

Our contributions are summarized as follows:
\begin{itemize}
    \item We demonstrate that a large language model can self-improve by taking datasets without ground truth outputs, by leveraging CoT reasoning~\citep{Wei2022ChainOT} and self-consistency~\citep{Wang2022SelfConsistencyIC}, achieving competitive in-domain multi-task performances as well as out-of-domain generalization. We achieve state-of-the-art-level results on ARC, OpenBookQA, and ANLI datasets.
    \item We provide detailed ablation studies on training sample formatting and sampling temperature after fine-tuning, and identify critical design choices for most successful self-improvement by LLMs.
    \item We study two other approaches for self-improvements, where the model generates additional questions from finite input questions and generates few-shot CoT prompt templates itself. The latter achieves 74.2\% on GSM8K, which is the state-of-the-art \textit{zero-shot} performance, against 43.0\% by~\citet{step_by_step} or 70.1\% through its naive extension with~\citet{Wang2022SelfConsistencyIC}. 
\end{itemize}


\input{Figures/framework}

%% file: Figures/framework.tex
\begin{figure*}[t]
\centering
\includegraphics[width=1.0\textwidth]{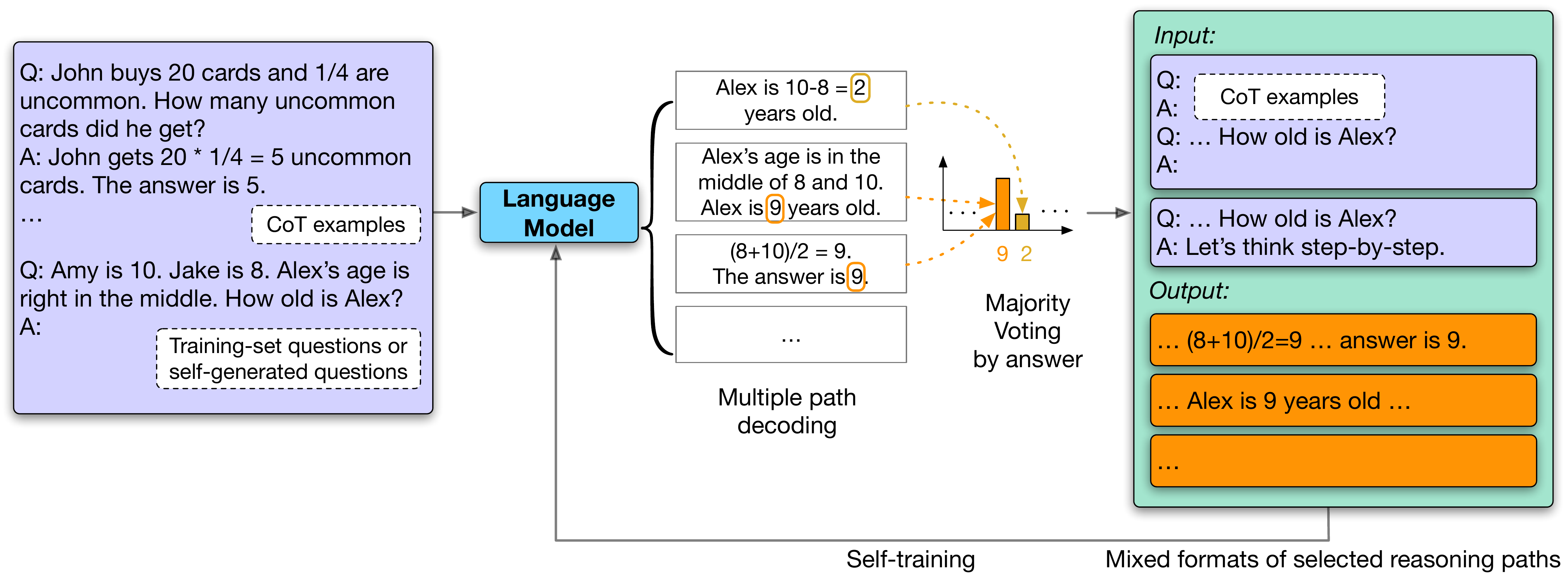}
\vspace{0.2em}
\caption{Overview of our method. With Chain-of-Thought (CoT) examples as demonstration~\citep{Wei2022ChainOT}, the language model generates multiple CoT reasoning paths and answers (temperature $T>0$) for each question. The most consistent answer is selected by majority voting~\citep{Wang2022SelfConsistencyIC}. The ``high-confidence'' CoT reasoning paths that lead to the majority answer are augmented by mixed formats as the final training samples to be fed back to the model for fine-tuning.
}
\label{fig:framework}
\end{figure*}

%% file: Sections/2-related.tex
\section{Related Work}
\label{sec:related_work}
\paragraph{Learning from explanations.}
Augmenting a machine learning model with explanations has been studied in existing literature extensively.
For example, in the supervised learning setting, a model can be fine-tuned using human-annotated rationales \citep{zaidan-etal-2007-using,ling-etal-2017-program,wt5,esnli,Cobbe2021TrainingVT,chung2022scaling}.
A few works have also looked at how explanations can help the models in various settings, e.g., in-context learning \citep{learn_from_explanation_in_context} and in distillation \citep{evaluate_explanations_distill}.
In this paper, we focus more on the \textit{unsupervised learning} setting, where we do not assume we have a rationale-augmented training dataset available, since human-annotated rationales can be expensive.

\paragraph{Few-shot explanations improves reasoning in LLMs.}
Recently, a lot of progress has been made towards improving LLMs' reasoning abilities via prompting or in-context learning.
\citet{Wei2022ChainOT} propose Chain-of-Thought prompting, which prompts the language model to generate a series of natural-language-based intermediate steps, and show it can help language models better solve complex and multi-step reasoning tasks.
\citet{Wang2022SelfConsistencyIC} improve Chain-of-Thought prompting by sampling multiple diverse reasoning paths and finding the most consistent answers via majority voting.
\citet{step_by_step} propose to prompt the language model with ``Let's think step by step'' to generate reasoning in a zero-shot fashion.
\citet{Zhou2022LeasttoMostPE} further decompose the questions into multiple sub-questions, and ask the language model to solve each sub-question sequentially.

\paragraph{Refining explanations.}
More recent work proposes to further refine the generated reasoning paths as some of them could be unreliable. For example, \citet{ye2022unreliability} calibrate model predictions based on the reliability of the explanations, \citet{maieutic_prompting} show that inducing a tree of explanations and inferring the satisfiability of each explanation can further help judge the correctness of explanations.
\citet{better_reasoner} show that sampling a diverse set of prompts from the training data, and a voting verifier can be used to improve model's reasoning performance.
\citet{star} proposes better rationale generation by augmenting ground truth answers as hints when predicted answers are incorrect. 
Our work is orthogonal to these lines of work, as we utilize refined explanations from~\citet{Wang2022SelfConsistencyIC} for fine-tuning the model for self-improvement, and could readily incorporate these other refinement techniques for generating higher-quality self-training data. Our work is similar to \citet{star} where we both propose to fine-tune a model on self-generated CoT data, but our method does not require ground truth labels and shows stronger empirical results with multi-task generalization.

\paragraph{Self-training models.}
One related line of work is self-training (see a survey from \citet{self-training-survey}). The key idea is to assign pseudo labels from a learned classifier to unlabeled data, and use these pseudo-labeled examples to further improve the original model training, e.g., \citep{DBLP:conf/cvpr/RoyChowdhuryCSJ19,self_training_xie,He2020Revisiting,chen2021-CPS}.
Different from such prior work, our proposed self-improvement framework uses CoT prompting plus self-consistency to obtain high-confidence solutions on a large set of unlabeled data to augment the fine-tuning process. 

\paragraph{Distillation and dark knowledge.} Our method also tangentially relates to rich literature on distillation~\citep{ba2014deep,hinton2015distilling}, where a student network imitates a teacher network's classifier predictions on input examples. A key detail is to learn from soft targets instead of hard predicted labels, as softmax outputs with a high temperature reveal more detailed relative class likelihoods, colloquially known as \textit{dark knowledge}~\citep{hinton2015distilling,korattikara2015bayesian}. Recent studies~\citep{star,snell2022learning,eisenstein2022honest} show that \textit{dark knowledge} within LLMs can be retrieved with more computation at inference time, such as adding informative instructions into the input sequence, and output CoT generation~\citep{Wei2022ChainOT,step_by_step}. In our work, we explicitly show that imperfect CoT reasoning (which may lead to incorrect answer) can be used directly for self-improving language models as evidenced in our experiments in Sections~\ref{sec:exp_limits} and~\ref{sec:exp_distill}.

%% file: Sections/3-method.tex
\section{Method}
\label{sec:method}

The overview of our method is illustrated in Fig.~\ref{fig:framework}:
We are given a pre-trained Large Language Model (LLM) $M$ and a question-only training dataset $\Dcal^{\mathtt{train}}=\{x_i\}_{i=1}^D$ with few-shot Chain-of-Thought (CoT) examples~\citep{Wei2022ChainOT}.
We apply multiple path decoding with a sampling temperature $T>0$ for generating $m$ reasoning paths and answers $\{r_{i_1}, r_{i_2}, \dots, r_{i_m}\}$ for each question $x_i$ in $\Dcal^{ \mathtt{train}}$, and use majority voting (self-consistency) to select the most consistent, highest confidence answer~\citep{Wang2022SelfConsistencyIC}. We then keep all reasoning paths that lead to the most consistent answer,  apply mixed formats of prompts and answers for augmentation, and fine-tune the model on these self-generated reasoning-answer data. We consider our approach as making the model self-improve. 
In the following sections, we detail important designs within our method, along with additional approaches for the model to self-improve without supervised data.

\subsection{Generating and Filtering Multiple Reasoning Paths}
Self-consistency~\citep{Wang2022SelfConsistencyIC} brings large improvements on reasoning tasks (e.g., $56.5\%\rightarrow74.4\%$ on GSM8K test set), and the gap between greedy decoding and diverse decoding shows there is a potential for further improving the reasoning ability of $M$, using the self-selected high-confidence reasoning paths as training data.
\input{Tables/self_consist}

For each training question ${x_i}$, we sample $m$ CoT reasoning paths, denoted as $\{r_{i_1}, r_{i_2}, \dots, r_{i_m}\}$ (see Table~\ref{tab:self_consist} for examples). 
Since $M$ is prompted with the CoT examples from~\citet{Wei2022ChainOT}, we apply the same output parsing with ``The answer is'' to generate their predicted answers  $\{y_{i_1}, y_{i_2}, \dots, y_{i_m}\}$.
The most consistent answer, which is not necessarily a correct answer, is selected by majority voting, denoted as $\tilde{y}_i=\argmax_{y_{i_j}}\sum_{k=1}^m\mathbb{I}(y_{i_j}=y_{i_k})$.
For all the training questions, we filter the CoT reasoning paths that reach $\tilde{y}$ as the final answer to be put into the self-training data, denoted as $\Dcal^{\mathtt{self-consistent}}=\{x_i, \tilde{\rvec_i}\}$, where $\tilde{\rvec_i} = \{r_{i_j}|1\leq j \leq m, y_{i_j} = \tilde{y}_i\}$.

\input{Figures/confidence_scatter}

Since we do not use any ground truth labels to filter out cases where $\tilde{y}_i\neq y_i$, it is important that the self-generated CoT reasoning paths are mostly reliable and incorrect answers do not hurt the self-improvement of the model.
We plot the relation between the accuracy and confidence of self-generated CoT paths for each question in GSM8K training set in Fig.~\ref{fig:scatter}. 
The confidence is the number of CoT paths leading to $\tilde{y}$ divided by the total path number $m$. The y-axis shows the accuracy of $\tilde{y}$ under a certain confidence. The circle area and the color darkness shows the number of questions under a certain confidence. 
We can observe that confident answers are more likely to be correct, which means that when a question has many consistent CoT paths, then the corresponding $\tilde{y}$ is more likely to be correct. On the other hand, when $\tilde{y}$ is wrong, it is likely to be supported by fewer CoT paths, and brings little noise to the training samples.


\subsection{Training with Mixed Formats}\label{sec:mix_formats}
To prevent the language model from overfitting to specific prompts or answer styles, we create four different formats for each reasoning path to be mixed in the self-training data, shown in Table~\ref{tab:mix_format}. In the first format, a few Chain-of-Thought examples (questions followed by reasoning paths leading to the correct final answers) are prepended to the new question, while the language model output is trained to be the same with the filtered CoT reasoning paths. In the second format, we use examples of questions and their direct answers as standard prompting, and the language model output is supposed to also only contain the direct answer. The third and fourth format are similar to the first and second format, except that no example of question-answer pairs are given, so that the model will learn to think on its own in an in-context zero-shot manner. In the third format, where we want the model to output CoT reasoning without prepending examples containing CoT reasonings, we append ``Let's think step by step.'' at the end of the input sequence, to guide the language model to generate step-by-step CoT reasoning paths \citep{step_by_step}. The mixed formats of training samples are then used to fine-tune the pre-trained language model $M$.
\input{Tables/mix_format}

\subsection{Generating Questions and Prompts}\label{sec:q_gen}
Given a set of training questions and a few human-written Chain-of-Thought (CoT) examples as prompts, our proposed approach enables model self-improvement. However, when the amount of training questions or CoT examples is limited, our method may not generate sufficient training samples for language model self-training. Collecting questions from the web requires human engineering. 
To further reduce human effort, we investigate how to self-generate more training questions as well as example prompts. 

\paragraph{Question Generation}\label{sec:q_gen}
Previous work~\citep{Yoo2021GPT3MixLL, Meng2022GeneratingTD} discuss few-shot data augmentation by generating diverse training samples using LLMs. However, those methods are designed for classification tasks and require ground truth label for each few-shot example. We use a simple yet effective approach to generate diverse questions (without ground truth answers) for in-domain questions. Specifically, we randomly select several existing questions, concatenate them in a random order as input prompt, and let the language model generate consecutive sequences as new questions. We repeat the process to obtain a large set of new questions, then use self-consistency~\citep{Wang2022SelfConsistencyIC} to only keep the questions that have a highly confident answer. Those questions are then used as self-generated training questions.

\paragraph{Prompt Generation}
Given a set of questions, humans can write CoT examples as reasoning paths leading to the final answer. In zero-shot setting without manual prompts, we can generate these CoT paths using the model itself. Following~\cite{step_by_step}, we start the answer with ``A: Let's think step by step.'' and let the language model generate the consecutive reasoning paths. We then use those generated reasoning paths as examples for few-shot CoT prompting.

%% file: Tables/self_consist.tex
\newcommand{\wronglabel}[1]{\textcolor{red}{(\textit{#1})}}
\begin{table}[h]
\caption{Examples of 3 self-generated CoT reasoning paths given a question. Output 1 and 3 are the most consistent reasoning paths based on majority voting and kept as self-training data.}
\centering
\resizebox{\textwidth}{!}{
\begin{tabular}{l}
\toprule
\textbf{Question}: Stefan goes to a restaurant with his family. They order an appetizer that costs \$10 and\\ 4 entrees that are \$20 each. If they tip 20\% of the total, what is the total amount of money that\\ they spend?\\
\midrule
\textbf{Multiple Path Decoding}:\\
\textbf{Output 1}: The appetizer costs \$10. The entrees cost \$20 each so in total \textcolor{blue}{4 * \$20 = \$80}. This means\\ the total cost is \textcolor{blue}{\$10 + \$80 = \$90}. They tip 20\% of it, so the total amount they spend is \textcolor{blue}{\$90 * 1.2}\\ \textcolor{blue}{= \$108}. The answer is 108.\\
\\
\textbf{Output 2}: The appetizer costs \$10 and the entrees are \$20 each. There are 4 entrees so the sum is\\ \$20 * 4 = \$80. The waiter gets 20\% of the total. 20\% of \$80 is \$80 * .2 = \$16. The answer is \$80\\ + \$16 = \$96. \wronglabel{Incorrect reasoning path}\\
\\
\textbf{Output 3}: The appetizer costs \$10. The entrees cost \textcolor{blue}{4 * \$20 = \$80}. The tip is 20\% of the total, so\\ it is 20\% of the \$90 they have spent. The tip is \textcolor{blue}{0.2 * 90 = \$18}. The total they spent is \textcolor{blue}{\$90 + \$18}\\ \textcolor{blue}{= \$108}. The answer is 108.\\
\bottomrule
\end{tabular}
}
\label{tab:self_consist}
\end{table}

%% file: Figures/confidence_scatter.tex
\begin{wrapfigure}{wr}{0.4\textwidth}
\vspace{-1.5em}
\begin{center}
    \includegraphics[width=\textwidth]{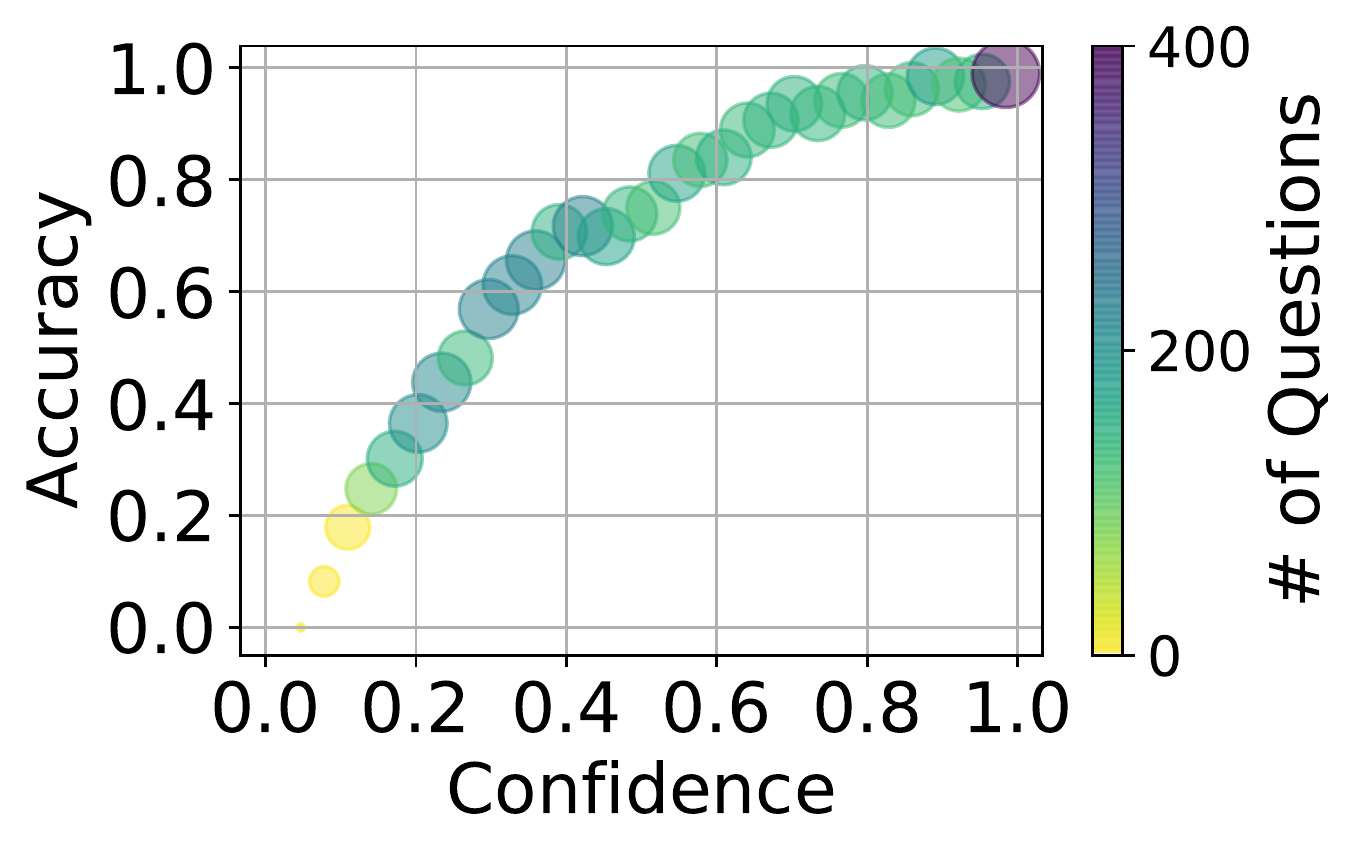}
\end{center}
\vspace{-0.4cm}
\caption{The relation of accuracy and confidence of the majority-voted answer after multiple path decoding on GSM8K training-set questions. Predicted confidence from self-consistency~\citep{Wang2022SelfConsistencyIC} is well calibrated~\citep{guo2017calibration}.}
\vspace{-0.3em}
\label{fig:scatter}
\end{wrapfigure}

%% file: Tables/mix_format.tex
\begin{table}[h]
\caption{An example of how a reasoning path is augmented into four formats of training data with different prompts (in input) and answer styles (in output). Specifically, the \textit{CoT prompting examples} used for each tasks are listed in Appendix~\ref{sec:app_prompt}. The \textit{Standard prompting examples} are the same question-answer pairs with \textit{CoT prompting examples}, except that reasoning is removed.}
\centering
\resizebox{\textwidth}{!}{
\begin{tabular}{l}
\toprule
\textbf{Question}: Amy is 10 years old. Jake is 8 years old. Alex's age is right in the middle. How old is Alex?\\
\textbf{Selected Chain-of-Thought}: Amy is 10 years old. Jake is 8 years old. Alex's age is in the middle of\\ Amy and Jake, so Alex is ( 8 + 10 ) / 2 = 9 years old. The answer is 9.\\
\midrule
\textbf{Mixed-formats of training data}:\\
\textbf{Format 1}: \textbf{Input}: \textit{[CoT prompting examples]} + `\textbackslash n' + \textit{[Question]} +  `\textbackslash n' + `A:'\\
\textbf{Output}: Amy is 10 years old. Jake is 8 years old. Alex's age is in the middle of Amy and Jake, so Alex\\ is ( 8 + 10 ) / 2 = 9 years old. The answer is 9.\\
\\
\textbf{Format 2}: \textbf{Input}: \textit{[Standard prompting examples]} + `\textbackslash n' + \textit{[Question]} +  `\textbackslash n' + `A:'\\
\textbf{Output}: The answer is 9.\\
\\
\textbf{Format 3}: \textbf{Input}: \textit{[Question]} +  `\textbackslash n' + `A: Let's think step by step.'\\
\textbf{Output}: Amy is 10 years old. Jake is 8 years old. Alex's age is in the middle of Amy and Jake, so Alex\\ is ( 8 + 10 ) / 2 = 9 years old. The answer is 9.\\
\\
\textbf{Format 4}: \textbf{Input}: \textit{[Question]} +  `\textbackslash n' + `A:'\\
\textbf{Output}: The answer is 9.\\
\bottomrule
\end{tabular}
}
\label{tab:mix_format}
\end{table}

%% file: Sections/4-setup.tex
\section{Experimental Setup}\label{sec:setup}

\paragraph{Tasks and Datasets.} 
We demonstrate the effectiveness of our method on three types of tasks\footnote{We evaluate on the test sets of GSM8K, ARC, OpenBookQA, and ANLI, and the dev set of DROP (ground truth labels of the test set are not publicly available).}:
\begin{itemize}[leftmargin=*]
\item \textbf{Arithmetic reasoning}: We use the math problem set GSM8K~\citep{Cobbe2021TrainingVT}, and a reading comprehension benchmark DROP~\citep{Dua2019DROPAR} which requires numerical reasoning. We follow~\cite{Zhou2022LeasttoMostPE} to partition the DROP dataset into football related and non-football related subsets for training.
\item \textbf{Commonsense reasoning}: We use the OpenBookQA~\citep{Mihaylov2018CanAS} dataset, and the AI2 Reasoning Challenge (ARC)~\citep{Clark2018ThinkYH} dataset. Note that for ARC, we only use the Challenge sub-set (ARC-c) in our experiments. Both datasets contain multiple-choice questions.
\item \textbf{Natural Language Inference}: We use the Adversarial NLI (ANLI)~\citep{Mihaylov2018CanAS} subsets, ANLI-A2 and ANLI-A3, which are the more challenging subsets compared to ANLI-A1. These datasets contain pairs of sentences with relations of entailment, neutral, or contradiction. 
\end{itemize}

\paragraph{Models, Training settings and Hyperparameters}
We follow previous studies~\citep{Wei2022ChainOT, Wang2022SelfConsistencyIC} and conduct our experiments on an autoregressive Transformer-based language model with 540 billion parameters. The CoT examples for each dataset are listed in Appendix~\ref{sec:app_prompt}. 
We generate $m=32$ reasoning paths for each question in a training set. Since each reasoning path is augmented into four formats in Sec.~\ref{sec:mix_formats}, the final training samples are up to the size of $128\times |\Dcal^\mathtt{train}|$, with $|\Dcal^\mathtt{train}|$ being the size of the corresponding training set. 
For all datasets except DROP, we use the whole training set; To reduce the training burden, we sample $5$k examples from the non-football and football partition of the DROP dataset, and sample $5$k examples from ANLI-A2 and ANLI-A3.
For each dataset, we fine-tune the model for 10k steps with a learning rate of $5$e$-5$ and a batch size of $32$.
For multiple path decoding, we use a sampling temperature of $T=0.7$ with the pre-trained model as suggested by~\cite{Wang2022SelfConsistencyIC}. We use $T=1.2$ for the language model after self-improvement (\method). We set the maximum number of decoded steps to $256$ for all experiments.

%% file: Sections/5-exp.tex
\section{Results}
\label{sec:results}
We conduct a series of experiments to demonstrate the effectiveness of our proposed self-improving method. First, we apply our method on each individual dataset (task) and report the results. We then merge the generated data from all datasets and train one model to study the generalization ability of the model on unseen datasets as in~\citep{wei2021finetuned}. In addition to the results of using generated CoT reasoning paths, we show studies on generating input questions and few-shot prompts. We end with ablation studies on model sizes and hyperparameters.

\subsection{Main Results}
\input{Tables/main_res}
We list the results of using the \bigmodel~model before and after \method in Table~\ref{tab:main_res}. For each model, during test time, we apply three separate prompting methods on all six datasets: standard-prompting, CoT-Prompting, and Self-Consistency. We observe that after \method, the performance of all three prompting methods increase by a large margin. We observe significant improvement, comparing self-consistency versus \method with self-consistency: $+7.7\%$ on GSM8K, $+4.8\%$ on DROP, $+4.4\%$ on OpenBookQA, and $+4.5\%$ on ANLI-A3.
This shows that our proposed method is quite effective. 
Furthermore, the single path CoT-Prompting performance of \method is close to or even better than the multiple path Self-Consistency performance of the model without \method, showing that \method truly helps the language model learn from the multiple consistent reasoning paths. 
We also compare our results with previous SOTA, achieved by different methods on different datasets, listed in Table~\ref{tab:main_res}. On ARC-c, OpenBookQA, ANLI-A2 and ANLI-A3, \method outperforms previous SOTA. On GSM8K dataset, \method is close to the DiVeRSe approach~\citep{Li2022OnTA} which uses diverse prompts and a voting verifier to ensemble $100$ output paths. On the contrary, we only use $32$ output paths for self-generating training samples and for self-consistency with \method. On the DROP dataset, \method is close to the OPERA approach~\citep{Zhou2022OPERAOD} which uses ground truth labels for training. On the other hand, our method only leverages the questions in the training set, without using ground truth labels.

\input{Tables/multitask_res}
\paragraph{Multi-task self-training for unseen tasks}
To demonstrate the generalization ability of \method, we conduct experiments of self-training on a mixture of the training-set questions from the above six datasets (denoted as In-Domain tasks), then use the same model checkpoint for the evaluation on six Out-Of-Domain (OOD) tasks, as shown in Table~\ref{tab:multi_res}. Of all the OOD tasks: (1) \textbf{AQUA}~\citep{Ling2017ProgramIB} and \textbf{SVAMP}~\citep{Patel2021AreNM} are arithmetic reasoning tasks; (2) \textbf{StrategyQA}~\citep{Geva2021DidAU} is a commonsense reasoning task; (3) \textbf{ANLI-A1}~\citep{Mihaylov2018CanAS}, \textbf{RTE}~\citep{Dagan2005ThePR} and \textbf{MNLI-M/MM}~\citep{Williams2018ABC} are natural language inference tasks.\footnote{We evaluate on the test set of SVAMP and ANLI, the dev set of MNLI and RTE (ground truth labels of the test sets are not publicly available). For StrategyQA we use the question-only set from~\cite{Srivastava2022BeyondTI}.} Among these tasks, \textbf{AQUA}, \textbf{StrategyQA}, and \textbf{RTE} are significantly different from any In-Domain task. These three tasks have their own few-shot prompts. From Table~\ref{tab:multi_res}, we can observe that \method achieves higher accuracy results on all OOD tasks, showing that the overall reasoning ability of the language model is improved.

\input{Tables/std_finetune}
\paragraph{Importance of training with Chain-of-Thought formats}
We demonstrate the importance of training language models with Chain-of-Thoughts compared to training with only direct answers. In Table~\ref{tab:std_ablate}, we list the results of \method with all four formats, and the results of \method with only direct answer formats.
The results clearly show that without the CoT formats, the language model can still self-improve, but the performance gain drops by a large amount compared to using all four formats.

\subsection{Pushing the limit of self-improvements}
\label{sec:exp_limits}

\input{Tables/q_gen}
\paragraph{Self-Generating Questions}
We further explore the few-shot setting where there are only limited training questions in the target domain. On GSM8K, we sample $10$ real questions as few-shot samples, and use the language model to generate more training questions using the method in Section~\ref{sec:q_gen}. We then self-train the language model with these generated questions and list the results in Table~\ref{tab:q_gen}. The results show that using self-generated questions still improves the reasoning ability of language models, but using the real training-set questions leads to better results.

\paragraph{Self-Generating Few-Shot CoT Prompts}
\input{Figures/zero_shot}
We explore the situation where no in-domain CoT examples are provided for a task. We apply the Step-by-Step method~\citep{step_by_step} to generate CoT examples using the language model as described in Section~\ref{sec:q_gen}, and show the results in Figure~\ref{fig:zero_shot}. We observe that few-shot prompting with self-generated Step-by-Step CoT examples substantially outperforms the Step-by-Step~\citep{step_by_step} baseline (66.2\% vs 53.8\% at 10 paths, 74.2\% vs 70.1\% at 40 paths), and nearly matches the performance of human-written few-shot CoT~\citep{wei2021finetuned} (74.4\% at 40 paths~\citep{Wang2022SelfConsistencyIC}). The strong performance of ``Few-Shot w/ Step-by-Step'' despite the limited accuracy of prompt examples (43.0\% for greedy Step-by-Step) likely comes from leveraging more diverse CoT prompts for multi-path decoding~\citep{Li2022OnTA}, where at 40 paths it uses 20 generate prompt-templates, each with 4-shot CoT examples, i.e. a total of 80 generated CoT examples compared to 8 human-written examples use in~\citet{Wei2022ChainOT}. Since we did not use training questions or few-shot CoT examples, 74.2\% also marks the new state-of-the-art zero-shot performance on GSM8K.


\subsection{Distillation to smaller models}
\label{sec:exp_distill}
\input{Tables/distill}
We also explore whether the knowledge can be distilled to smaller models, such as in distillation~\citep{hinton2015distilling} and in~\citet{star}. We use the same set of training samples generated by the \bigmodel~model, but fine-tune on models with smaller sizes (\smallmodel~and \midmodel~respectively), and show the results of CoT-prompting in Table~\ref{tab:distill}. It is interesting to point out that after distillation from \method, the 62 billion model can outperform the pre-trained 540 billion model, and the 8 billion model can outperform the pre-trained 62 billion model. This implies that for downstream applications with limited computing resources, the reasoning knowledge from large models can be used to largely enhance small models to achieve competitive performance.

\subsection{Hyperparameter Study}
\input{Figures/hyper_study}
\paragraph{Sampling Temperature after Self-Improvement}
We study the effect of varying the temperature $T$ for multiple path decoding after \method is applied. Specifically, we vary $T$ between [$0.7, 1.0, 1.2, 1.5$] and show the results on GSM8K and DROP dataset respectively in Fig.~\ref{fig:temp_study}. As shown in the figure, $T=1.2$ benefits both datasets the most, and is used in the Self-Consistency method for \method on all datasets. We notice that the optimal $T$ after model self-improvement is larger than the optimal $T=0.7$~\citep{Wang2022SelfConsistencyIC} before self-improvement. We believe the reason is that after training the model, the entropy of the output distribution is reduced.

\paragraph{Number of Sampled Reasoning Paths}
We study whether the number of sampled reasoning paths $m$ for Self-Consistency largely affects the accuracy after \method is applied. We show the accuracy on GSM8K test set for models both with or without \method in Fig.~\ref{fig:path_study}. For both cases, setting $m=15$ already achieves a reasonably good accuracy, and using a larger $m$ only brings marginal improvements. We also notice that after Self-Improvement, using $5$ paths for Self-Consistency can already surpass the performance of using $32$ paths for model without Self-Improvement. Thus, with a well-improved model, huge computing resources can be saved when applied to real applications.

%% file: Tables/main_res.tex
\begin{table}[h]
\caption{Accuracy results on six reasoning benchmarks. The previous SOTA results are from: (a) \cite{Li2022OnTA}, (b) \cite{Zhou2022OPERAOD}, (c) \cite{Wang2022SelfConsistencyIC}, (d) \cite{Wang2022RationaleAugmentedEI}.}
\vspace{1.5em}
\centering
\resizebox{\textwidth}{!}{
\begin{tabular}{l*{8}{c}}
\toprule
& \textbf{Prompting Method}
& \textbf{GSM8K} & \textbf{DROP} & \textbf{ARC}-c & \textbf{OpenBookQA} & \textbf{ANLI}-A2 & \textbf{ANLI}-A3 \\ 
\midrule
& Previous SOTA & 82.3$^\text{\textit{a}}$ & 84.9$^\text{\textit{b}}$ & 88.7$^\text{\textit{c}}$ & 91.0$^\text{\textit{d}}$ & 64.9$^\text{\textit{d}}$ & 66.0$^\text{\textit{d}}$\\
\midrule
\multirow{3}{*}{w/o \method} &
Standard-Prompting  & 17.9 & 60.0 & 87.1 & 84.4 & 55.8 & 55.8 \\
& CoT-Prompting & 56.5 & 70.6 & 85.2 & 86.4 & 58.9 &  60.6 \\
& Self-Consistency & 74.4 & 78.2 & 88.7 & 90.0 & 64.5 & 63.4\\
\midrule
\multirow{3}{*}{\method} &
Standard-Prompting  & 32.2 & 71.7 & 87.2 & 92.0 & 64.8 & 66.9  \\
& CoT-Prompting & 73.5 & 76.2 & 88.3 & 93.0  & 65.3 & 67.3 \\
& Self-Consistency  & 82.1 & 83.0 & \textbf{89.8} & \textbf{94.4} & \textbf{66.5} & \textbf{67.9} \\ 
\bottomrule
\end{tabular}
}
\label{tab:main_res}
\end{table}

%% file: Tables/multitask_res.tex
\begin{table}[h]
\caption{Comparison of CoT-prompting accuracy results on six Out-Of-Domain benchmarks with or without training on six In-Domain (GSM8K, DROP, ARC-c, OpenBookQA, ANLI-A2, ANLI-A3) training-set questions.}
\vspace{1.5em}
\centering
\resizebox{\textwidth}{!}{
\begin{tabular}{l*{8}{c}}
\toprule
& \textbf{Self-training data}
& \textbf{AQUA} & \textbf{SVAMP} & \textbf{StrategyQA} & \textbf{ANLI}-A1 & \textbf{RTE} & \textbf{MNLI}-M/MM \\ 
\midrule
\multirow{1}{*}{w/o \method} 
& - & 35.8 & 79.0 & 75.3 & 68.8 & 79.1 & 72.0/74.0  \\
\midrule
\multirow{1}{*}{\method} 
& GSM8K + DROP + ... & 39.0 & 82.8 & 77.8 & 79.2 &  80.1  &  81.8/82.2 \\
\bottomrule
\end{tabular}
}
\label{tab:multi_res}
\end{table}

%% file: Tables/std_finetune.tex
\begin{table}[h]
\caption{Ablation study: w/ or w/o CoT reasoning paths as training format on GSM8K dataset.}
\vspace{1.5em}
\centering
\resizebox{0.66\textwidth}{!}{
\begin{tabular}{l*{3}{c}}
\toprule
& \multicolumn{2}{c}{Results on \textbf{GSM8K}} \\ 
& Standard Prompting & CoT Prompting \\ 
\midrule
w/o \method & 17.9 & 56.5 \\
\midrule
\method w/o CoT formats & 23.6 & 61.6 \\
\midrule
\method & 32.2 & 73.5 \\
\bottomrule
\end{tabular}
}
\label{tab:std_ablate}
\end{table}

%% file: Tables/q_gen.tex
\begin{table}[h]
\caption{Accuracy on GSM8K test set after self-training on self-generated or training set questions.}
\vspace{1.5em}
\centering
\resizebox{0.72\textwidth}{!}{
\begin{tabular}{l*{4}{c}}
\toprule
& \textbf{Questions used}
& \multicolumn{2}{c}{Results on \textbf{GSM8K}} \\ 
 & \textbf{for Self-Training} & CoT-Prompting & Self-Consistency \\
\midrule
w/o \method & -
& 56.5 & 74.4 \\
\midrule
\method & Generated Questions & 66.2 & 78.1\\ 
\midrule
\method & Training-set Questions & \textbf{73.5} & \textbf{82.1}\\ 
\bottomrule
\end{tabular}
}
\label{tab:q_gen}
\end{table}

%% file: Figures/zero_shot.tex
\begin{figure}[h]
\centering
\includegraphics[width=0.45\textwidth]{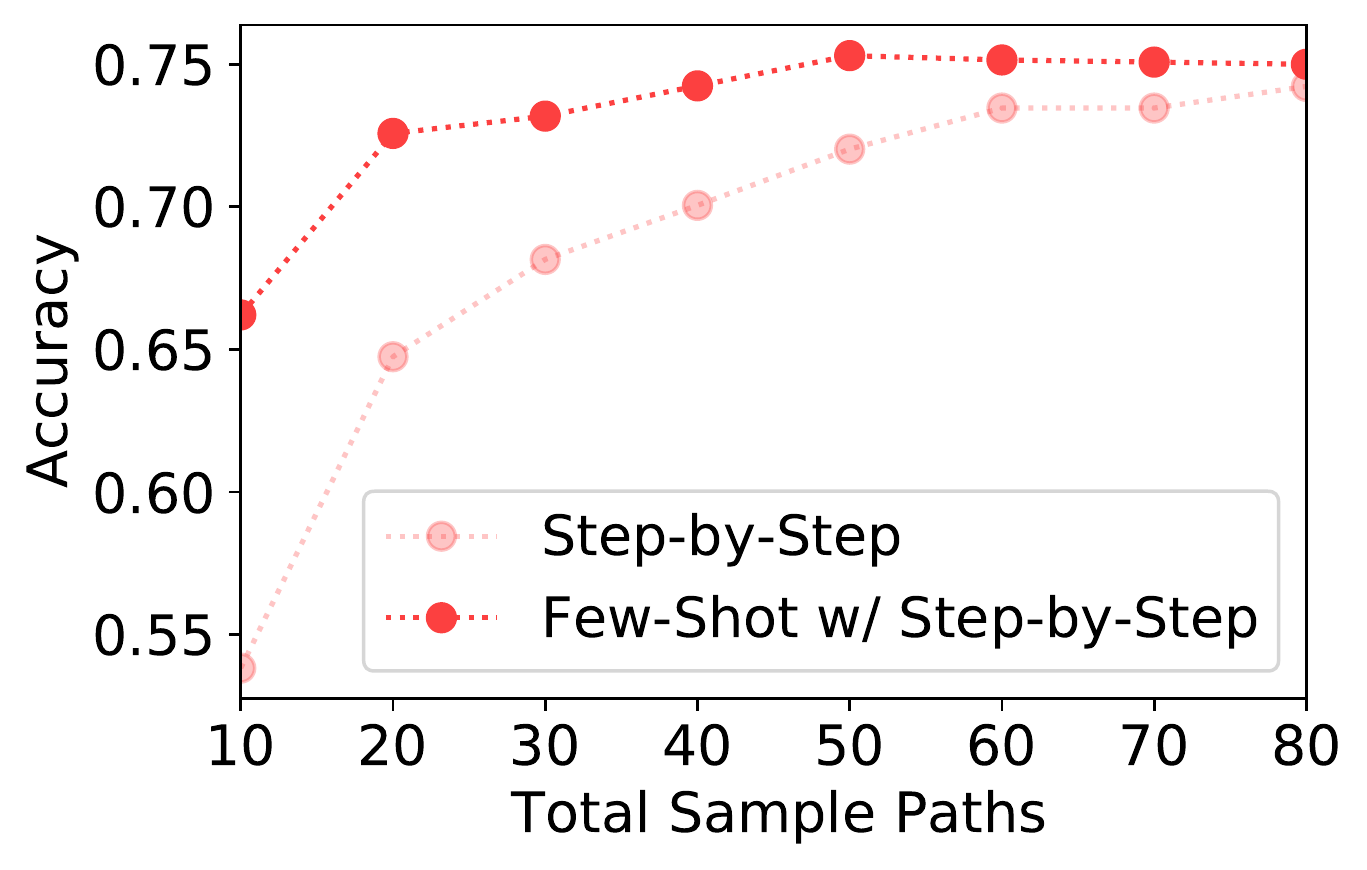}
\vspace{0.2em}
\caption{Accuracy results on GSM8K test set using \bigmodel~model with multi-path sampling and self-consistency~\citep{Wang2022SelfConsistencyIC}. ``Step-by-Step'' is the baseline performance of ~\citet{step_by_step} plus self-consistency~\citep{Wang2022SelfConsistencyIC}, while our ``Few-Shot w/ Step-by-Step'' uses exemplers self-generated from Step-by-Step (greedy decoding) for few-shot prompting the LLM.
}
\label{fig:zero_shot}
\end{figure}



%% file: Tables/distill.tex
\begin{table}[h]
\caption{Distillation from \bigmodel~model to small models. We see that distilled smaller models outperform models that are one-tier larger.}
\centering
\resizebox{0.7\textwidth}{!}{
\begin{tabular}{l*{3}{c}}
\toprule
& \multicolumn{3}{c}{Results on \textbf{GSM8K}} \\
& 8 billion & 62 billion & 540 billion \\
\midrule
w/o \method & 5.0 & 29.7 & 56.5 \\
\midrule
Distilled from \method 540 billion & 33.4 & 57.4 & - \\
\bottomrule
\end{tabular}
}
\label{tab:distill}
\end{table}

%% file: Figures/hyper_study.tex
\begin{figure}[h]
\subfigcapmargin=10pt
\centering
\subfigure[Accuracy results of \method on GSM8K and DROP test set when different sampling temperatures are applied for Self-Consistency.]{
\label{fig:temp_study}
\includegraphics[width=0.35\textwidth]{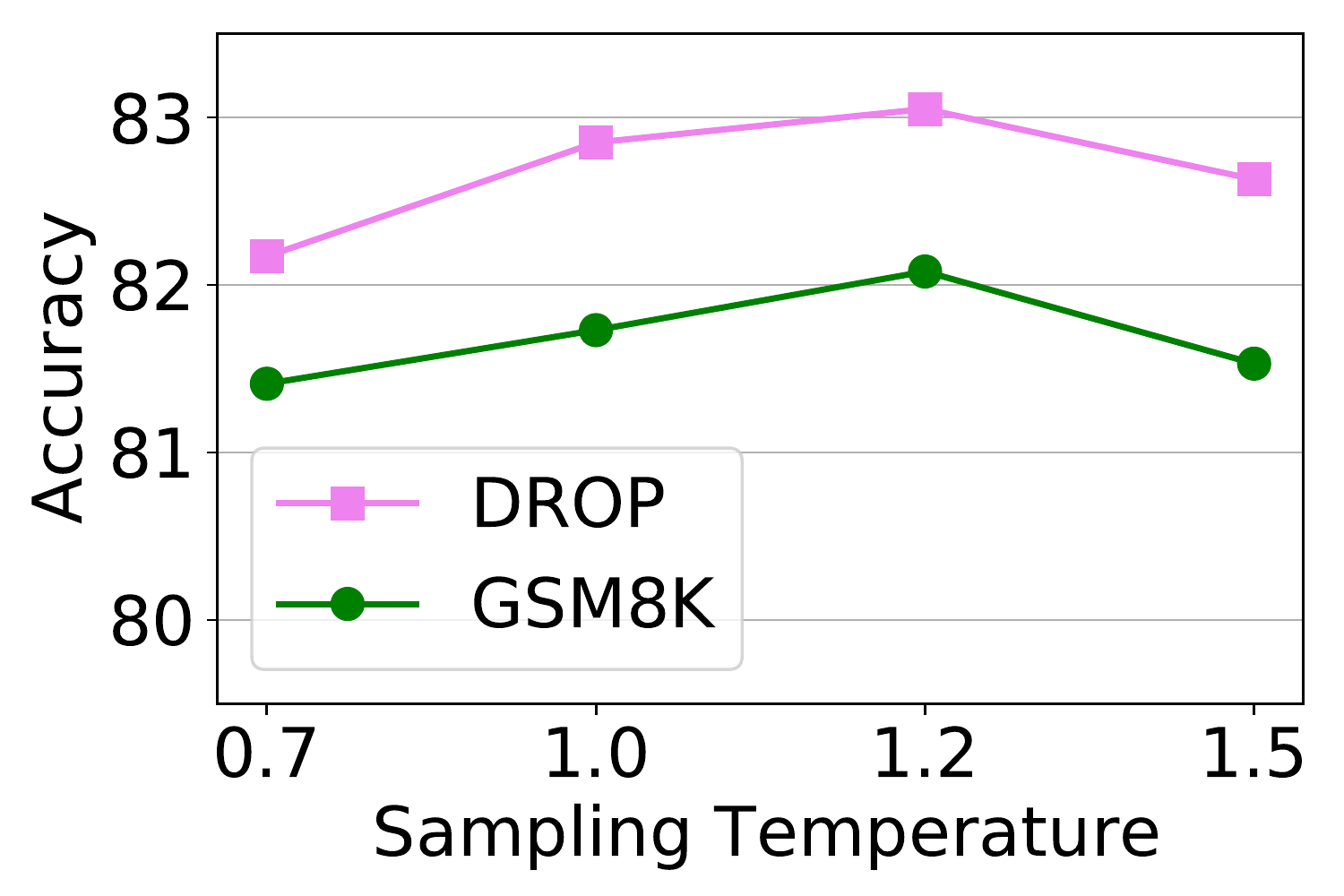}
\vspace{0.2em}
}
\subfigure[Accuracy results with or without \method on GSM8K test set using different numbers of sampled reasoning path for Self-Consistency.]{
\label{fig:path_study}
\includegraphics[width=0.35\textwidth]{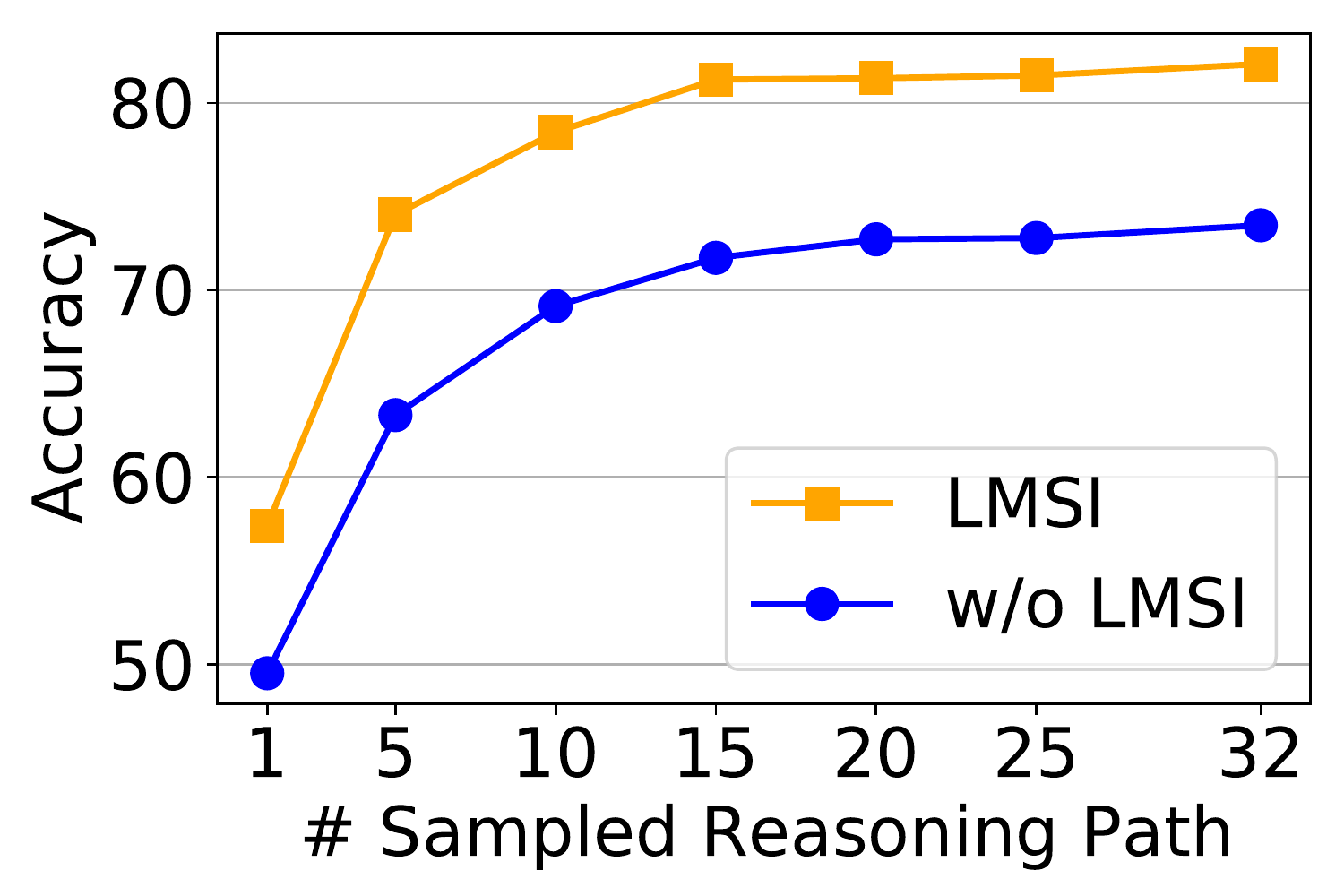}
\vspace{0.2em}
}
\vspace{-0.2cm}
\caption{Hyperparameter study results.}\label{fig:hyper_study}
\vspace{-0.2em}
\end{figure}

%% file: Sections/6-concl.tex
\section{Conclusions}
\label{sec:conclusions}

We demonstrated that a Large Language Model (LLM) is capable of improving its performance on reasoning datasets by training on its own generated labels, given input questions only. Experiments using an LLM with 540 billion parameters show that our approach improves the accuracy scores on the six datasets by 1.1\% to 7.7\%, achieving new state-of-the-art results on ARC, OpenBookQA, and ANLI, without training on ground truth labels. Furthermore, we show that it is possible for the LLM to self-improve even on its own generated questions and few-shot Chain-of-Thought prompts. As part of our future work, we plan to combine large-scale generated data from our approach and existing supervised data, to further improve the performance of LLMs.

%% file: Sections/ack.tex
\section*{Acknowledgements}
\label{sec:acknowledgements}
We thank Jason Wei (Google research), Hyung Won Chung (Google research), and Denny Zhou (Google research) for their advice and feedback on our work. We thank Yi Tay (Google research) for the guidance on UL2 fine-tuning. We also thank Jingjin Li (Cornell University) for the discussion about metacognition.

%% file: Sections/7-app.tex
\clearpage
\appendix
\section{Appendix}
\label{app:appendix}

\subsection{Results on UL2 model}
We also apply \method on a recently proposed public language model, UL2~\citep{Tay2022UL2UL}, using the pre-trained model at step 2,650,000\footnote{UL2: \url{https://github.com/google-research/google-research/tree/master/ul2}}. We use a fixed set of hyperparameters for fine-tuning on each dataset. Specifically, we generate $m=40$ reasoning paths for each question in a training set for majority voting. We fine-tune the model for 10k steps with a learning rate of $5$e$-5$ and a batch size of $32$. For multiple path decoding, we use a sampling temperature of $T=0.5$ with the pre-trained UL2 model following~\citet{Tay2022UL2UL}, and set $T=0.7$ for the language model after \method. We set the maximum number of decode steps to $256$ for all experiments.

The results are shown in Table~\ref{tab:ul2_res}. For arithmetic reasoning datasets, we follow~\citep{Tay2022UL2UL} to provide both exact matching accuracy scores as well as accuracy scores after an equation-correction postprocessing step. We observe that for most datasets, \method still improves the reasoning accuracy, but the improvement on UL2 is not as large as that on \bigmodel. We think the reason is that, since \method exploits the implicit rationale of language models, and the capacity of a language model is determined by its size, larger models can capture more high-order semantics and are more likely to benefit from \method.

\input{Tables/ul2_res}

\subsection{Chain-of-Thought Prompts for Each Dataset}~\label{sec:app_prompt}
We list the Chain-of-Thought Prompts for each dataset for ``CoT-Prompting'' experiments and self-generated training samples.

\input{Tables/gsm8k_prompt}

%% file: Tables/ul2_res.tex
\begin{table}[h]
\caption{Accuracy results on six reasoning benchmarks with \method on UL2. On GSM8K and DROP, we also include accuracy scores after an equation-correction postprocessing step.}
\vspace{1.5em}
\centering
\resizebox{\textwidth}{!}{
\begin{tabular}{l*{8}{c}}
\toprule
& \textbf{Prompting Method}
& \textbf{GSM8K} & \textbf{DROP} & \textbf{ARC}-c & \textbf{OpenBookQA} & \textbf{ANLI}-A2 & \textbf{ANLI}-A3 \\ 
\midrule
\multirow{2}{*}{w/o \method} 
& CoT-Prompting & 5.4/7.1  & 11.1/16.8 & 49.9 & 53.6 & 35.9 & 33.8 \\
& Self-Consistency & 6.4/9.9 & 16.8/26.5 & 54.9  & 54.0 & 37.4 & 36.8\\
\midrule
\multirow{2}{*}{\method} 
& CoT-Prompting & 6.1/8.6 & 11.4/17.1 & 50.9 & 53.8 & 35.4 & 34.4 \\
& Self-Consistency  & 7.9/10.2 & 18.1/28.1 & 54.7 & 55.2 & 38.1 & 37.4 \\ 
\bottomrule
\end{tabular}
}
\label{tab:ul2_res}
\end{table}

%% file: Tables/gsm8k_prompt.tex
\begin{table}[h]
    \caption{
    Few-shot CoT prompts for OpenBookQA, from~\cite{Wang2022RationaleAugmentedEI}.
    }
    \centering
    \small
    \begin{tabular}{p{14cm}}
        \toprule
        \textbf{Q:} Poison causes harm to which of the following? (a) a Tree (b) a robot (c) a house (d) a car\\
        \vspace{-1mm}
        \textbf{A:} Poison will harm living things, only a tree is a living thing. The answer is (a).\\
        \vspace{0mm}
        \textbf{Q:} As you look deeper into a Marbel you can see (a) the future (b) minut defects (c) colors (d) the other side\\
        \vspace{-1mm}
        \textbf{A:} Marbel is not transparent, so you can not see the other side. Marbel does not necessarily have multiple colors. You will see minut defects. The answer is (b). \\
        \vspace{0mm}
        \textbf{Q:} When food is reduced in the stomach (a) the mind needs time to digest (b) take a second to digest what I said (c) nutrients are being deconstructed (d) reader's digest is a body of works\\
        \vspace{-1mm}
        \textbf{A:} The food is being deconstructed in the stomach during digestion. The answer is (c).\\
        \vspace{0mm}
        \textbf{Q:} The sun is responsible for (a) puppies learning new tricks (b) children growing up and getting old (c) flowers wilting in a vase (d) plants sprouting, blooming and wilting\\
        \vspace{-1mm}
        \textbf{A:} The sun can affect the growing of living things, like plants. The answer is (d).\\
        \bottomrule
    \end{tabular}
    \label{tab:prompt-obqa}
\end{table}

\begin{table*}[h]
    \caption{
    Few-shot CoT prompts for GSM8K and SVAMP, from~\cite{Wei2022ChainOT}.
    }
    \centering
    \small
    \begin{tabular}{p{13.5cm}}
        \toprule
        \textbf{Q:} There are 15 trees in the grove. Grove workers will plant trees in the grove today. After they are done, there will be 21 trees. How many trees did the grove workers plant today? \\
        \vspace{-1mm}
        \textbf{A:} We start with 15 trees. Later we have 21 trees. The difference must be the number of trees they planted. So, they must have planted 21 - 15 = 6 trees. The answer is 6. \\
        \vspace{0mm}
        \textbf{Q:} If there are 3 cars in the parking lot and 2 more cars arrive, how many cars are in the parking lot? \\
        \vspace{-1mm}
        \textbf{A:} There are 3 cars in the parking lot already. 2 more arrive. Now there are 3 + 2 = 5 cars. The answer is 5. \\
        \vspace{0mm}
        \textbf{Q:} Leah had 32 chocolates and her sister had 42. If they ate 35, how many pieces do they have left in total? \\
        \vspace{-1mm}
        \textbf{A:} Leah had 32 chocolates and Leah's sister had 42. That means there were originally 32 + 42 = 74 chocolates. 35 have been eaten. So in total they still have 74 - 35 = 39 chocolates. The answer is 39. \\
        \vspace{0mm}
        \textbf{Q:} Jason had 20 lollipops. He gave Denny some lollipops. Now Jason has 12 lollipops. How many lollipops did Jason give to Denny? \\
        \vspace{-1mm}
        \textbf{A:} Jason had 20 lollipops. Since he only has 12 now, he must have given the rest to Denny. The number of lollipops he has given to Denny must have been 20 - 12 = 8 lollipops. The answer is 8. \\
        \vspace{0mm}
        \textbf{Q:} Shawn has five toys. For Christmas, he got two toys each from his mom and dad. How many toys does he have now? \\
        \vspace{-1mm}
        \textbf{A:} He has 5 toys. He got 2 from mom, so after that he has 5 + 2 = 7 toys. Then he got 2 more from dad, so in total he has 7 + 2 = 9 toys. The answer is 9. \\
        \vspace{0mm}
        \textbf{Q:} There were nine computers in the server room. Five more computers were installed each day, from monday to thursday. How many computers are now in the server room? \\
        \vspace{-1mm}
        \textbf{A:} There are 4 days from monday to thursday. 5 computers were added each day. That means in total 4 * 5 = 20 computers were added. There were 9 computers in the beginning, so now there are 9 + 20 = 29 computers. The answer is 29. \\
        \vspace{0mm}
        \textbf{Q:} Michael had 58 golf balls. On tuesday, he lost 23 golf balls. On wednesday, he lost 2 more. How many golf balls did he have at the end of wednesday? \\
        \vspace{-1mm}
        \textbf{A:} Michael initially had 58 balls. He lost 23 on Tuesday, so after that he has 58 - 23 = 35 balls. On Wednesday he lost 2 more so now he has 35 - 2 = 33 balls. The answer is 33.\ \\
        \vspace{0mm}
        \textbf{Q:} Olivia has \$23. She bought five bagels for \$3 each. How much money does she have left? \\
        \vspace{-1mm}
        \textbf{A:} She bought 5 bagels for \$3 each. This means she spent 5 * \$3 = \$15 on the bagels. She had \$23 in beginning, so now she has \$23 - \$15 = \$8. The answer is 8. \\
        \bottomrule
    \end{tabular}
    \label{tab:prompt-math}
\end{table*}

\begin{table*}[h]
    \caption{
    Few-shot CoT prompts for DROP (nonfootball), from~\cite{Zhou2022LeasttoMostPE}.
    }
    \centering
    \small
    \begin{tabular}{p{13.5cm}}
        \toprule
        \textbf{Q:} Since the 1970s, U.S. governments have negotiated managed-trade agreements, such as the North American Free Trade Agreement in the 1990s, the Dominican Republic-Central America Free Trade Agreement in 2006, and a number of bilateral agreements. In Europe, six countries formed the European Coal and Steel Community in 1951 which became the European Economic Community in 1958. Two core objectives of the EEC were the development of a common market, subsequently renamed the single market, and establishing a customs union between its member states. How many years did the European Coal and Steel Community exist? \\
        \vspace{-1mm}
        \textbf{A:} According to the passage, the European Coal and Steel Community was established in 1951 and became the EEC in 1958. 1958 - 1951 = 7. So the answer is 7. \\
        \vspace{0mm}
        \textbf{Q:} In the county, the population was spread out with 23.50\% under the age of 18, 8.70\% from 18 to 24, 29.70\% from 25 to 44, 24.70\% from 45 to 64, and 13.30\% who were 65 years of age or older. How many more percent are under the age of 18 compared to the 18 to 24 group? \\
        \vspace{-1mm}
        \textbf{A:} According to the passage, 23.5\% are under the age of 18, and 8.7\% are from ages 18 to 24. 23.5\% - 8.7\% = 14.8\%. So the answer is 14.8. \\
        \vspace{0mm}
        \textbf{Q:} Playing in their second straight Thanksgiving game, the Eagles struggled especially on defense, where they were unable to stop the much-hyped Lions offense. The worst of it all was how unproven rookie Eric Rowe was tasked with covering wide receiver Calvin Johnson, leading to Johnson catching 3 touchdowns. Stafford's five passing touchdowns, including three of them to Johnson was too much for the Eagles to overcome and for the second consecutive time this season, the Eagles gave up 45 points in a game. With the loss, the Eagles drop to 4-7 on the season and 6-1 when playing on Thanksgiving. How many TD passes did Stafford throw other than to Johnson? \\
        \vspace{-1mm}
        \textbf{A:} According to the passage, Stafford threw 5 TD passes, 3 of which were to Johnson. 5 - 3 = 2. So the answer is 2. \\
        \bottomrule
    \end{tabular}
    \label{tab:prompt-dropnf}
\end{table*}

\begin{table*}[h]
    \caption{
    Few-shot CoT prompts for DROP (football), from~\cite{Zhou2022LeasttoMostPE}.
    }
    \centering
    \small
    \begin{tabular}{p{13.5cm}}
        \toprule
        \textbf{Q:} The Seahawks played the San Francisco 49ers. In the first quarter, the Hawks RB Julius Jones got a 27-yard TD run, along with DT Craig Terrill returning a fumble 9 yards for a touchdown. In the third quarter, the 49ers almost rallied as RB H. J. Torres made a 12-yard TD pass to Lucas Nelly, along with Mare kicking a 32-yard field goal. In the final quarter, Julius Jones got another 11-yard TD. How many yards do the shortest touchdown run and the longest touchdown pass combine for? \\
        \vspace{-1mm}
        \textbf{A:} All the touchdown runs are: a 27-yard touchdown run, a 9-yard touchdown run, a 11-yard touchdown run. The smallest number among 27, 9, 11 is 9. So the shortest touchdown run was 9 yards. All the touchdown passes are: a 12-yard touchdown pass. So the longest touchdown pass was 12 yards. So the shortest touchdown run and the longest touchdown pass combine for 9 + 12 = 21 yards. So the answer is 21 yards. \\
        \vspace{0mm}
        \textbf{Q:} The Steelers went home for a duel with the Baltimore Ravens. Pittsburgh would deliver the opening punch in the first quarter with a 1-yard touchdown from running back Rashard Mendenhall. The Ravens would make it even as running back Willis McGahee got a 9-yard TD. The Ravens kicker Billy Cundiff got a 45-yard field goal in the second quarter, concluding the first half with a 10-7 lead. The Steelers brought the game into overtime with a 38-yard field goal by Andrew Foster. The Ravens Billy Cundiff pulled off a winning 33-yard field goal in overtime. How many points did the Ravens have at halftime? \\
        \vspace{-1mm}
        \textbf{A:} The Ravens kicker Billy Cundiff got a 45-yard field goal in the second quarter, concluding the first half with a 10-7 lead.  So the Ravens had 10 points at halftime. So the answer is 10 points. \\
        \vspace{0mm}
        \textbf{Q:} The Vikings flew to Bank of America Stadium to face the Carolina Panthers. After a scoreless first quarter, Carolina got on the board with quarterback Matt Moore finding fullback Brad Hoover on a 1-yard TD pass. After yet another scoreless quarter, Carolina sealed the game as Matt Moore completed a 42-yard touchdown pass to wide receiver Steve Smith. How many scoreless quarters were there? \\
        \vspace{-1mm}
        \textbf{A:} The first and third quarters were the scoreless quarters. So there are 2 scoreless quarters. So the answer is 2. \\
        \bottomrule
    \end{tabular}
    \label{tab:prompt-dropf}
\end{table*}

\begin{table}[h]
    \caption{
    Few-shot CoT prompts for NLI tasks, including ANLI and MNLI, from~\cite{Wang2022RationaleAugmentedEI}.
    }
    \centering
    \small
    \begin{tabular}{p{14cm}}
        \toprule
        Premise:\\
        "Conceptually cream skimming has two basic dimensions - product and geography."\\
        Based on this premise, can we conclude the hypothesis "Product and geography are what make cream skimming work." is true?\\
        OPTIONS:\\
        - yes\\
        - no\\
        - it is not possible to tell\\
        \vspace{-1mm}
        \textbf{A:} Based on "cream skimming has two basic dimensions" we can't infer that these two dimensions are what make cream skimming work. The answer is it is not possible to tell.\\
        \vspace{0mm}
        Premise:\\
        "One of our member will carry out your instructions minutely."\\
        Based on this premise, can we conclude the hypothesis "A member of my team will execute your orders with immense precision." is true?\\
        OPTIONS:\\
        - yes\\
        - no\\
        - it is not possible to tell\\
        \vspace{-1mm}
        \textbf{A:} "one of" means the same as "a member of", "carry out" means the same as "execute", and "minutely" means the same as "immense precision". The answer is yes.\\
        \vspace{0mm}
        Premise:\\
        "Fun for adults and children."\\
        Based on this premise, can we conclude the hypothesis "Fun for only children." is true?\\
        OPTIONS:\\
        - yes\\
        - no\\
        - it is not possible to tell\\
        \vspace{-1mm}
        \textbf{A:} "adults and children" contradicts "only children". The answer is no.\\
        \vspace{0mm}
        Premise:\\
        "He turned and smiled at Vrenna."\\
        Based on this premise, can we conclude the hypothesis "He smiled at Vrenna who was walking slowly behind him with her mother." is true?\\
        OPTIONS:\\
        - yes\\
        - no\\
        - it is not possible to tell\\
        \vspace{-1mm}
        \textbf{A:} the premise does not say anything about "Vrenna was walking". The answer is it is not possible to tell.\\
        \vspace{0mm}
        Premise:\\
        "well you see that on television also"\\
        Based on this premise, can we conclude the hypothesis "You can see that on television, as well." is true?\\
        OPTIONS:\\
        - yes\\
        - no\\
        - it is not possible to tell\\
        \vspace{-1mm}
        \textbf{A:} "also" and "as well" mean the same thing. The answer is yes.\\
        \vspace{0mm}
        Premise:\\
        "Vrenna and I both fought him and he nearly took us."\\
        Based on this premise, can we conclude the hypothesis "Neither Vrenna nor myself have ever fought him." is true?\\
        OPTIONS:\\
        - yes\\
        - no\\
        - it is not possible to tell\\
        \vspace{-1mm}
        \textbf{A:} "Vrenna and I both" contradicts "neither Vrenna nor myself". The answer is no.\\
        \bottomrule
    \end{tabular}
    \label{tab:prompt-nli}
\end{table}

\begin{table}[h]
    \caption{
    Few-shot CoT prompts for ARC-challenge, from~\cite{Wang2022SelfConsistencyIC}.
    }
    \centering
    \small
    \begin{tabular}{p{13.5cm}}
        \toprule
        \textbf{Q:} George wants to warm his hands quickly by rubbing them. Which skin surface will produce the most heat? (a) dry palms. (b) wet palms. (c) palms covered with oil. (d) palms covered with lotion.\\
        \vspace{-1mm}
        \textbf{A:} Dry surfaces will more likely cause more friction via rubbing than other smoother surfaces, hence dry palms will produce the most heat. The answer is (a).\\
        \vspace{0mm}
        \textbf{Q:} Which factor will most likely cause a person to develop a fever? (a) a leg muscle relaxing after exercise. (b) a bacterial population in the bloodstream. (c) several viral particles on the skin. (d) carbohydrates being digested in the stomach.\\
        \vspace{-1mm}
        \textbf{A:} Option (b), bacterial population is the most likely cause for a person developing fever. The answer is (b). \\
        \vspace{0mm}
        \textbf{Q:} Which change in the state of water particles causes the particles to become arranged in a fixed position? (a) boiling. (b) melting. (c) freezing. (d) evaporating.\\
        \vspace{-1mm}
        \textbf{A:} When water is freezed, the particles are arranged in a fixed position; the particles are still moving for all other options. The answer is (c).\\
        \vspace{0mm}
        \textbf{Q:} When a switch is used in an electrical circuit, the switch can (a) cause the charge to build. (b) increase and decrease the voltage. (c) cause the current to change direction. (d) stop and start the flow of current.\\
        \vspace{-1mm}
        \textbf{A:} The function of a switch is to start and stop the flow of a current. The answer is (d).\\
        \bottomrule
    \end{tabular}
    \label{tab:prompt-arc}
\end{table}

\begin{table}[h]
    \caption{
    Few-shot CoT prompts for AQUA, from~\cite{Wang2022SelfConsistencyIC}.
    }
    \centering
    \small
    \begin{tabular}{p{13.5cm}}
        \toprule
        \textbf{Q:} John found that the average of 15 numbers is 40. If 10 is added to each number then the mean of the numbers is? Answer Choices: (a) 50 (b) 45 (c) 65 (d) 78 (e) 64\\
        \vspace{-1mm}
        \textbf{A:} If 10 is added to each number, then the mean of the numbers also increases by 10. So the new mean would be 50. The answer is (a). \\
        \vspace{0mm}
        \textbf{Q:} If a / b = 3/4 and 8a + 5b = 22,then find the value of a. Answer Choices: (a) 1/2 (b) 3/2 (c) 5/2 (d) 4/2 (e) 7/2\\
        \vspace{-1mm}
        \textbf{A:} If a / b = 3/4, then b = 4a / 3. So 8a + 5(4a / 3) = 22. This simplifies to 8a + 20a / 3 = 22, which means 44a / 3 = 22. So a is equal to 3/2. The answer is (b).\\
        \vspace{0mm}
        \textbf{Q:} A person is traveling at 20 km/hr and reached his destiny in 2.5 hr then find the distance? Answer Choices: (a) 53 km (b) 55 km (c) 52 km (d) 60 km (e) 50 km\\
        \vspace{-1mm}
        \textbf{A:} The distance that the person traveled would have been 20 km/hr * 2.5 hrs = 50 km. The answer is (e).\\
        \vspace{0mm}
        \textbf{Q:} How many keystrokes are needed to type the numbers from 1 to 500? Answer Choices: (a) 1156 (b) 1392 (c) 1480 (d) 1562 (e) 1788\\
        \vspace{-1mm}
        \textbf{A:} There are 9 one-digit numbers from 1 to 9. There are 90 two-digit numbers from 10 to 99. There are 401 three-digit numbers from 100 to 500. 9 + 90(2) + 401(3) = 1392. The answer is (b).\\
        \bottomrule
    \end{tabular}
    \label{tab:prompt-aqua}
\end{table}

\begin{table}[h]
    \caption{
    Few-shot CoT prompts for StrategyQA, from~\cite{Wei2022ChainOT}.
    }
    \centering
    \small
    \begin{tabular}{p{13.5cm}}
        \toprule
        \textbf{Q:} Do hamsters provide food for any animals?\\
        \vspace{-1mm}
        \textbf{A:} Hamsters are prey animals. Prey are food for predators. Thus, hamsters provide food for some animals. The answer is yes.\\
        \vspace{0mm}
        \textbf{Q:} Could Brooke Shields succeed at University of Pennsylvania?\\
        \vspace{-1mm}
        \textbf{A:} Brooke Shields went to Princeton University. Princeton University is about as academically rigorous as the University of Pennsylvania. Thus, Brooke Shields could also succeed at the University of Pennsylvania. The answer is yes.\\
        \vspace{0mm}
        \textbf{Q:} Yes or no: Hydrogen's atomic number squared exceeds number of Spice Girls?\\
        \vspace{-1mm}
        \textbf{A:} Hydrogen has an atomic number of 1. 1 squared is 1. There are 5 Spice Girls. Thus, Hydrogen's atomic number squared is less than 5. The answer is no.\\
        \vspace{0mm}
        \textbf{Q:} Yes or no: Is it common to see frost during some college commencements?\\
        \vspace{-1mm}
        \textbf{A:} College commencement ceremonies can happen in December, May, and June. December is in the winter, so there can be frost. Thus, there could be frost at some commencements. The answer is yes.\\
        \vspace{0mm}
        \textbf{Q:} Yes or no: Could a llama birth twice during War in Vietnam (1945-46)?\\
        \vspace{-1mm}
        \textbf{A:} The War in Vietnam was 6 months. The gestation period for a llama is 11 months, which is more than 6 months. Thus, a llama could not give birth twice during the War in Vietnam. The answer is no.\\
        \vspace{0mm}
        \textbf{Q:} Yes or no: Would a pear sink in water?\\
        \vspace{-1mm}
        \textbf{A:} The density of a pear is about 0.6 $\text{g/cm}^3$, which is less than water. Objects less dense than water float. Thus, a pear would float. The answer is no.\\
        \bottomrule
    \end{tabular}
    \label{tab:prompt-strategyqa}
\end{table}

\begin{table}[h]
    \caption{
    Few-shot CoT prompts for RTE, from~\cite{Wang2022RationaleAugmentedEI}.
    }
    \centering
    \small
    \begin{tabular}{p{14cm}}
        \toprule
        Premise:\\
        "No Weapons of Mass Destruction Found in Iraq Yet."\\
        Based on this premise, can we conclude the hypothesis "Weapons of Mass Destruction Found in Iraq." is true?\\
        \vspace{-1mm}
        \textbf{A:} "No Weapons of Mass Destruction Found" contradicts "Weapons of Mass Destruction Found". The answer is no.\\
        \vspace{0mm}
        Premise:\\
        "A place of sorrow, after Pope John Paul II died, became a place of celebration, as Roman Catholic faithful gathered in downtown Chicago to mark the installation of new Pope Benedict XVI."\\
        Based on this premise, can we conclude the hypothesis "Pope Benedict XVI is the new leader of the Roman Catholic Church." is true?'\\
        \vspace{-1mm}
        \textbf{A:} "installation of new Pope Benedict XVI." means "Pope Benedict XVI is the new leader". The answer is yes. \\
        \vspace{0mm}
        Premise:\\
        "A man is due in court later charged with the murder 26 years ago of a teenager whose case was the first to be featured on BBC One's Crimewatch. Colette Aram, 16, was walking to her boyfriend's house in Keyworth, Nottinghamshire, on 30 October 1983 when she disappeared. Her body was later found in a field close to her home. Paul Stewart Hutchinson, 50, has been charged with murder and is due before Nottingham magistrates later."\\
        Based on this premise, can we conclude the hypothesis "Paul Stewart Hutchinson is accused of having stabbed a girl." is true?\\
        \vspace{-1mm}
        \textbf{A:} The premise does not say Paul Stewart Hutchinson "stabbed" this girl. The answer is no.\\
        \vspace{0mm}
        Premise:\\
        "Herceptin was already approved to treat the sickest breast cancer patients, and the company said, Monday, it will discuss with federal regulators the possibility of prescribing the drug for more breast cancer patients."\\
        Based on this premise, can we conclude the hypothesis "Herceptin can be used to treat breast cancer." is true?\\
        \vspace{-1mm}
        \textbf{A:} "Herceptin was approved to treat breast cancer" implies that "Herceptin can be used to treat breast cancer". The answer is yes.\\
        \bottomrule
    \end{tabular}
    \label{tab:prompt-rte}
\end{table}